\immediate\write18{pdflatex \jobname}

\documentclass{styles/svproc}
\usepackage{url}

\usepackage[table]{xcolor}

\usepackage{graphicx} 
\usepackage{url}
\usepackage{hyperref}

\usepackage{pgfplots}
\usepackage{pgfplotstable}
\pgfplotsset{compat=1.17}

\usepackage{svg}
\svgpath{{graphs/}}

\definecolor{Gray}{gray}{0.9}

\usepackage{subcaption}

\begin{document}
\mainmatter              
\title{Sim-to-Real gap in RL: Use Case with TIAGo and Isaac Sim/Gym}
\titlerunning{Sim-to-Real gap in RL}  
%
\author{Jaume Albardaner \and Alberto San Miguel,
Néstor García \and Magí Dalmau}
\authorrunning{Jaume Albardaner et al.} 
%
\tocauthor{Jaume Albardaner, Alberto San Miguel, Néstor García, Magí Dalmau}
\institute{Eurecat, Centre Tecnològic Catalunya, Robotics \& Automation Unit, Barcelona, Spain\\
\email{\{jaume.albardaner, alberto.sanmiguel, nestor.garcia, magi.dalmau\}@eurecat.org}}

\maketitle              

\begin{abstract} 

This paper explores policy-learning approaches in the context of sim-to-real transfer for robotic manipulation using a TIAGo mobile manipulator, focusing on two state-of-art simulators, Isaac Gym and Isaac Sim, both developed by Nvidia. Control architectures are discussed, with a particular emphasis on achieving collision-less movement in both simulation and the real environment. Presented results demonstrate successful sim-to-real transfer, showcasing similar movements executed by an RL-trained model in both simulated and real setups.

\end{abstract}
\section{Reinforcement Learning simulators}

Reinforcement Learning (RL) techniques are especially useful for applications that require a certain degree of autonomy. In Robotics, this applies to tasks related to object manipulation and navigation, and the main difficulty lies on the gathering of data to train a model. An economically expensive approach to make it faster is to acquire multiple robots and run them simultaneously~\cite{multirobot1}~\cite{multirobot2}~\cite{multirobot3}. Models trained with that data are more likely to work on the real robot, since the data was obtained from the same platform the model runs on. However, performing RL on real robots still requires supervision to ensure that no unexpected scenario is met.This approach is not only expensive, it also requires a lot of time for data generation. Not only because of how slow the robot may move, but also because the environment may need to be prepared before every run.

Following this need of making the process faster and cheaper, several libraries for RL were developed such as OpenAI's Gymnasium~\cite{gymnasium} or Stable-baselines~\cite{stablebaselines} and robot models were introduced into simulators, like Mujoco~\cite{mujoco}. Although this process was definitely cheaper, it did not necessarily make the process faster. Following hardware development, simulators with GPU support appeared, among which both IsaacGym~\cite{isaacgym} and IsaacSim~\cite{isaacsim} from Nvidia stand out. Nonetheless, what happens inside the simulators is nothing more than a simplification of reality. Thereafter, an adaptation must be made to models if these are to be utilized in a real environment. This is a very active branch of RL called sim-to-real (or Sim2Real), which has been the main focus of this paper.

\section{TIAGo use case}



\subsection{Models}
Simulating physics for objects that don't have simple meshes is computationally expensive. If possible, simulators simplify these meshes into convex shapes to speed up the simulation. Additionally, in most simulators the default settings for robots include not checking for self collisions. 

Such is the case for the TIAGo mobile manipulator robot\footnote{\url{pal-robotics.com/robots/TiaGo/}}: the simulator settings had to be changed to allow self-collision and the model for the robot had to be modified because of the self-collisions caused by the mesh simplification.

\subsection{Controls}
The way the robot is controlled and how it reacts to control inputs differs in each simulator. In this section the control pipeline of the simulators that were used, as well as the real robot, are described.

\subsubsection{Isaac Gym}
\begin{figure}[ht]
    \centering
    \includegraphics[width=0.6\textwidth]{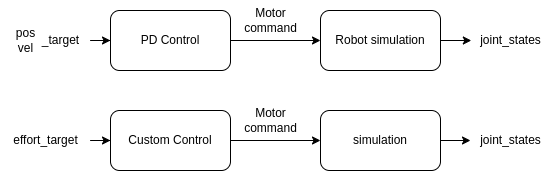}
    \caption{Control pipeline for Isaac Gym and Isaac Sim.}
    \label{fig:isaacGym-pipeline}
\end{figure}

Three types of joint controls can be applied: position, velocity and effort controls. As described in Figure \ref{fig:isaacGym-pipeline}, when the reference value for the position or velocity is updated, a PD controller is used to track the new reference. The PD controller takes its $k_p$ and $k_d$ values from the \textit{stiffness} and \textit{damping} parameters of each joint. The resulting control is delivered to the robot, which ensures no joint limitations are violated and simulates its effects on each joint.

In Isaac Gym there is no built-in controller for effort control. However, sample controllers are provided in Nvidia's IsaacGymEnvs GitHub repository\footnote{github.com/NVIDIA-Omniverse/IsaacGymEnvs}. 

\subsubsection{Isaac Sim}
Since this simulator also relies on PhysX and has also been developed by Nvidia, the control pipeline that is followed is the same as Figure \ref{fig:isaacGym-pipeline}. The only difference is that a wide variety of controllers are provided off-the-shelf. These controllers can operate the robot in joint and cartesian space.

\subsubsection{TIAGo}
\begin{figure}[ht]
    \centering
    \includegraphics[width=0.8\linewidth]{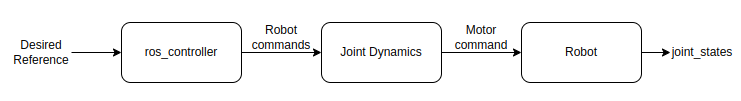}
    \caption{Control pipeline for the real TIAGo.}
    \label{fig:real-control}
\end{figure}

The robot that has been used in this use case follows the control schematics established by the \verb|ros_control| Robot Operating System (ROS) package\footnote{\url{github.com/ros-controls/ros_control}} (Figure \ref{fig:real-control}). Since it is community driven each step of the process can be customized and evaluated independently.

In the case of TIAGo, the \emph{JointGroupPosition} controller has been employed. This controller was selected because it's the most similar one to Isaac's PD: it is a PID.
\section{Results} \vspace{-0.1cm}
\subsection{Simulator responses}
 In order to evaluate the difference in responses in each simulator with the real robot, a step input was introduced to each joint of the arm. Although the individual response for each joint was similar in both simulators, when steps were introduced simultaneously to several joints responses differed. 

\begin{figure}[h!]
    \centering
    \begin{subfigure}{0.47\textwidth}
        \includegraphics[width=\textwidth]{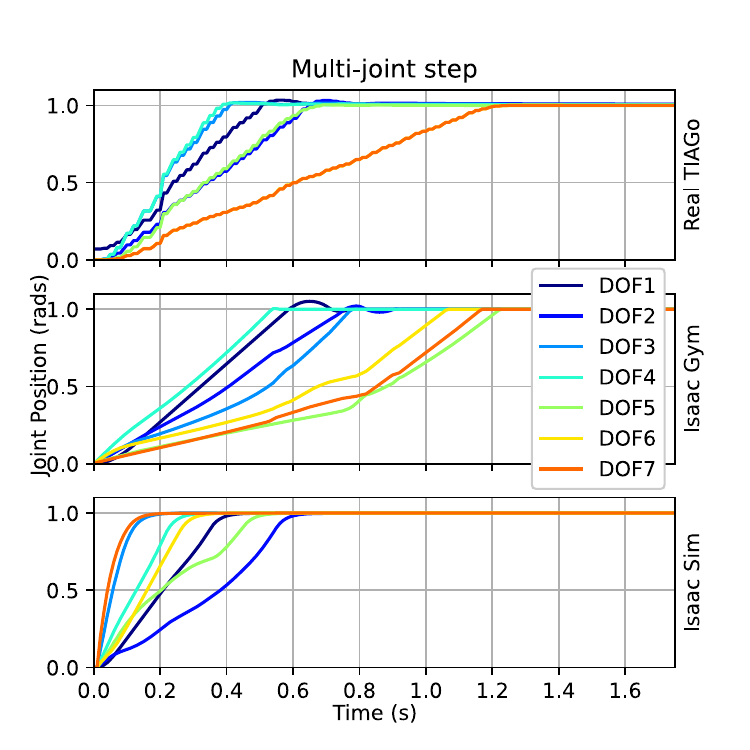} 
    \end{subfigure}
    \begin{subfigure}{0.47\textwidth}
        \includegraphics[width=\textwidth]{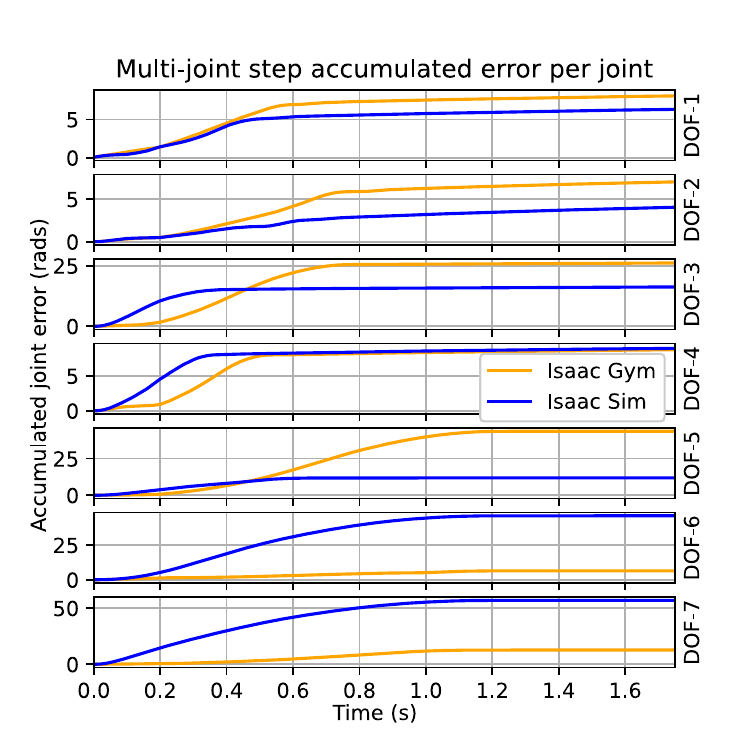} 
    \end{subfigure}
    \caption{Responses for each DOF in TIAGo's arm to simultaneous step inputs on each joint in the real TIAGo, Isaac Gym and Isaac Sim (a) and the accumulated errors of each simulator w.r.t the real (b)}
    \label{fig:stepResponse}
\end{figure}

\begin{figure}[h!]
    \centering
    \includegraphics[width=0.73\textwidth]{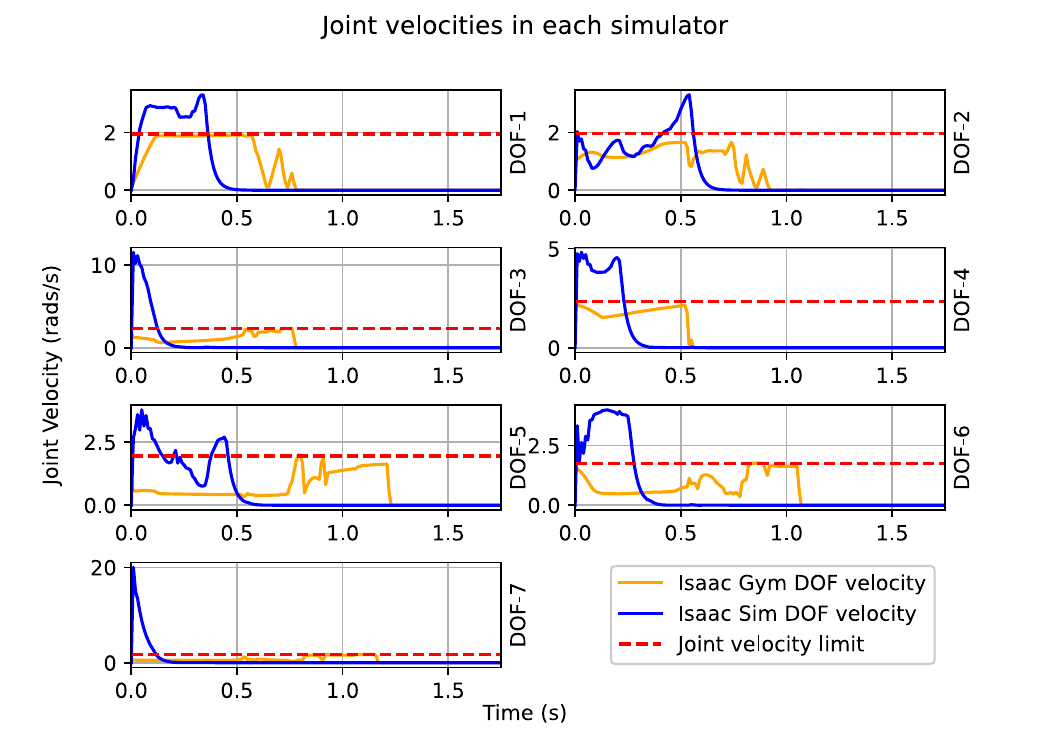} 
    \caption{Velocities for each DOF in TIAGo's arm to simultaneous step inputs on each joint in Isaac Gym and Isaac Sim}
    \label{fig:stepResponse_vel}
\end{figure}


Figure \ref{fig:stepResponse} shows that responses for the arm in Isaac Gym resemble better the results obtained in the real TIAGo than those obtained in Isaac Sim. Nonetheless, it seems to have a slower response, possibly due to prioritizing each joint based on their position in the hyerarchical tree. This hypothesis is based on the evolution of the accumulated errors, where errors seem to converge slower on DOF 7 than in DOF 1. Additionally, accumulated errors are also smaller for Isaac Gym than for Isaac Sim, as seen in Table \ref{tab:accumulated}. On the other hand, the difference in response with respect to Isaac Sim is most likely due to velocity constraints not being respected at a joint level, as seen in Figure \ref{fig:stepResponse_vel}.

\begin{table}[h!]
    \centering
    \begin{tabular}{|c|c|c|c|c|c|c|c|c|} \hline \rowcolor{Gray}
         & \textbf{DOF 1} & \textbf{DOF 2} & \textbf{DOF 3} & \textbf{DOF 4} & \textbf{DOF 5} & \textbf{DOF 6} & \textbf{DOF 7} & $\sum$\\ \hline
        Isaac Gym & 8.4202 & 7.5179 & 26.4569 & \textbf{9.0018} & 43.6909 & \textbf{6.46288} & \textbf{12.7499} & \textbf{114.3006}\\
        Isaac Sim & \textbf{6.6786 }& \textbf{4.6277} & \textbf{16.5728} & 9.1713 & \textbf{11.9117} & 46.1073 & 57.0088 & 152.0785\\ \hline
    \end{tabular}
    \vspace{1mm}
    \caption{Accumulated errors (rad) for each DOF after steady state has been achieved. The lowest values are highlighted in bold.}
    \label{tab:accumulated}
\end{table}


\vspace{-2mm}

\subsection{Trained model}

\begin{figure}[ht]
    \centering
    \includegraphics[width=0.45\textwidth]{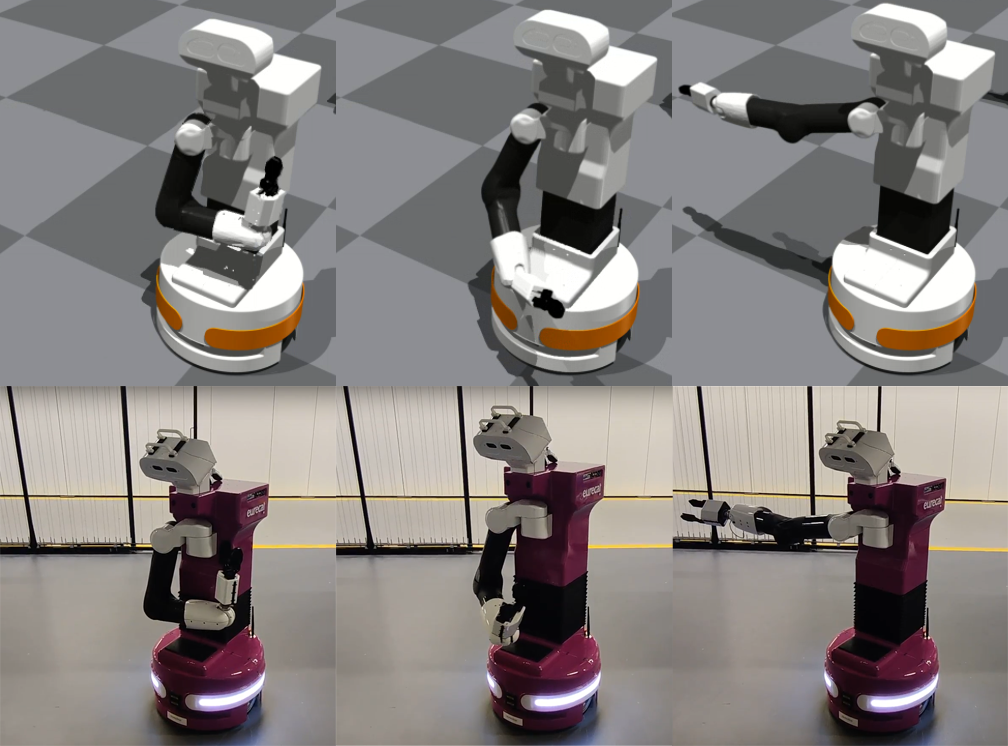}
    \includegraphics[width=0.5\textwidth]{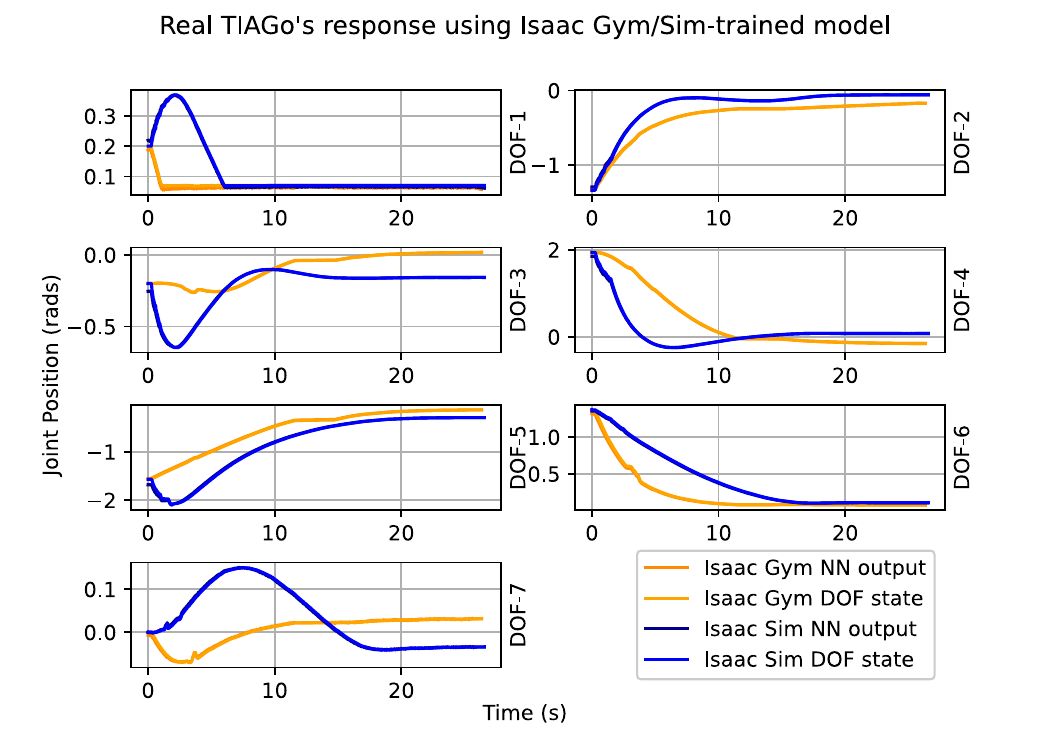}     
    \caption{Home-to-zero movement from the TIAGo in Isaac Gym (left-top), in the real setup (left-bottom) and the DOF positions through time for the models trained in Isaac Gym and Isaac Sim (right).}
    \label{fig:home2zero}
\end{figure}

The first model that was trained with this setup takes the TIAGo mobile manipulator from its ``Home'' position to a position where the arm is fully extended (Figure \ref{fig:home2zero}), which corresponds to all joints being in the zero position.

In Figure \ref{fig:home2zero} it can be seen that although both models were trained using the same reward function (reward$= -|DOF_{value}|$) for the same number of training epochs (100K), the movement differs.


\section{Conclusions}

In this paper, policy learning approaches were brought to the context of sim-to-real transfer for robotic platforms, particularly using Isaac Gym and Isaac Sim simulators for the TIAGo mobile manipulator. While presented results showcase promising achievements, identified issues should be addressed in the future to reduce their sim-to-real gap for this case and evaluate their effect on other robotic platforms.

%
%

\end{document}